\documentclass{bmvc2k}
\usepackage{hyperref}

\title{Metric Learning for \\ Novelty and Anomaly Detection}

\addauthor{Marc Masana}{mmasana@cvc.uab.cat}{1}
\addauthor{Idoia Ruiz}{iruiz@cvc.uab.cat}{1}
\addauthor{Joan Serrat}{joans@cvc.uab.cat}{1}
\addauthor{Joost van de Weijer}{joost@cvc.uab.cat}{1}
\addauthor{Antonio M. Lopez}{antonio@cvc.uab.cat}{1}

\addinstitution{
 Computer Vision Center\\
 Universitat Aut\`onoma de Barcelona\\
 Bellaterra, Spain
}

\runninghead{M. Masana et al.}{Metric Learning for Novelty and Anomaly Detection}

\usepackage{multirow}

\newcommand{\minisection}[1]{\vspace{0.04in} \noindent {\bf #1}\ }
\begin{document}

\maketitle
\vspace{-0.25cm}
\begin{abstract}
When neural networks process images which do not resemble the distribution seen during training, so called out-of-distribution images, they often make wrong predictions, and do so too confidently. The capability to detect out-of-distribution images is therefore crucial for many real-world applications. We divide out-of-distribution detection between novelty detection ---images of classes which are not in the training set but are related to those---, and anomaly detection ---images with classes which are unrelated to the training set. By related we mean they contain the same type of objects, like digits in MNIST and SVHN. Most existing work has focused on anomaly detection, and has addressed this problem considering networks trained with the cross-entropy loss. Differently from them, we propose to use metric learning which does not have the drawback of the softmax layer (inherent to cross-entropy methods), which forces the network to divide its prediction power over the learned classes. We perform extensive experiments and evaluate both novelty and anomaly detection, even in a relevant application such as traffic sign recognition, obtaining comparable or better results than previous works.
\end{abstract}

\section{Introduction}
\label{sec:intro}

Deep neural networks have obtained excellent performance for many applications. However, one of the known shortcomings of these systems is that they can be overly confident when presented with images (and classes) which were not present in the training set. Therefore, a desirable property of these systems would be the capacity to not produce an answer if an input sample belongs to an unknown class, that is, a class for which it has not been trained. The field of research which is dedicated to this goal is called out-of-distribution detection~\cite{hendrycks-gimpel,liang-li,lee-lee}. Performing out-of-distribution detection is important not only to avoid classification errors but also as the first step towards lifelong learning systems~\cite{chen2016lifelong}. Such systems would detect out-of-distribution samples in order to later update the model accordingly~\cite{kirkpatrick2017overcoming,liu2018rotate}.

The problem of out-of-distribution detection has also been called one-class classification, novelty and anomaly detection~\cite{pimentel-clifton}. More recently, associated to deep neural network classifiers, some works refer to it as open-set recognition~\cite{bendale-boult16}. In this paper, we distinguish two cases of out-of-distribution which we believe are quite different: we propose to term as \textit{novelty} an image from a class different from those contained in a dataset from which to train, but that bears some resemblance to them, for instance because it shows the same kind of object from untrained points of view. This is a very important problem in many computer vision applications.
For example, imagine a system that classifies traffic signs on-board a car and takes automatic decisions accordingly. It can happen that it finds a class of local traffic signs which was not included in the training set, and this must be detected to avoid taking wrong decisions. We reserve the word \textit{anomaly} for completely unrelated samples, like different type of objects, images from another unrelated dataset, or background patches in the case of traffic sign classification. This is also relevant from the point of view of commercial applications. In fact, most previous works focus on anomaly detection. Novelty detection remains rather unexplored. To the best of our knowledge only~\cite{schultheiss-kading} and~\cite{liang-li} perform some intra-dataset out-of-distribution detection experiments. The three previous works closest to ours~\cite{hendrycks-gimpel,liang-li,lee-lee}, revolve around one idea: given a discriminative neural network model, use the output probabilities to take the decision of seen/unseen class. These networks are optimized to distinguish between the classes present in the training set, and are not required to explicitly model the marginal data distribution. As a consequence, at testing time the system cannot assess the probability of the presented data, complicating the assessment of novelty cases.

Here we explore a completely different approach: to learn an embedding where one can use Euclidean distance as a measure of ``out-of-distributioness''. We propose a loss that learns an embedding where samples from the same in--distribution class form clusters, well separated from the space of other in--distribution classes {\em and} also from out-of-distribution samples. The contributions to the problem of out-of-distribution detection presented in this paper are the following. First, the use of metric learning for out-of-distribution detection, instead of doing it on the basis of the cross-entropy loss and corresponding softmax scores. Second, we distinguish between novelty and anomaly detection and show that research should focus on the more challenging problem of novelty detection. Third, we obtain comparable or better results than state-of-the-art in both anomaly and novelty detection. Last, in addition to the experiments with benchmark datasets in order to compare with previous works, we address also a real-world classification problem, traffic sign recognition, for which we obtain good detection {\em and} accuracy results.

\section{Related work}
Our paper is related to anomaly detection in its different meanings. Also to open-set recognition, as one of the most important applications of out-of-distribution detection. And finally to metric learning, the base of our approach. In the following we briefly review the most related works in each of these areas. Out-of-distribution detection should not be confused with another desirable property of machine learning systems, namely the reject option, that is, the ability to decide not to classify an input if the confidence on any of the labels is too weak (see for example~\cite{geifman-yaniv} and references therein). The difference is that in the latter case it is assumed that the sample does belong to some class present during training.
\smallskip

\minisection{Anomaly and novelty detection.} Also known as out-of-distribution detection, it aims at identifying inputs that are completely different from or unknown to the original data distribution used for training~\cite{pimentel-clifton}. In~\cite{Bodesheim13}, they perform novelty detection by learning a distance in an embedding. It proposes a Kernel Null Foley-Sammon transform that aims at projecting all the samples of each in-distribution class into a single point in a certain space. Consequently, novelty detection can be performed by thresholding the distance of a test sample to the nearest of the collapsed class representations. However, they employ handcrafted features, thus optimizing only the transform parameters and not the representation, like in the presently dominating paradigm of deep learning.

Although Deep Neural Networks (DNNs) have established as state-of-the-art on many computer vision classification and detection tasks, overconfidence in the probability score of such networks is a common problem. DNNs capable of detecting lots of objects with fine accuracy can still be fooled by predicting new never-seen objects with high confidence. This problem can be defined by the ability of the network to decide if a new test sample belongs to the in-distribution (i.e. from a class or from the data used to train the classifier) or to an out-of-distribution.

In~\cite{hendrycks-gimpel}, they show that DNNs trained on MNIST~\cite{lecun1998gradient} images can frequently produce high confidence guesses (+90\%) on random noise images. They propose a baseline for evaluation of out-of-distribution detection methods and show that there is room for future research to improve that baseline. Their baseline assumes that out-of-distribution samples will have a more distributed confidence among the different classes than an in-distribution sample. Recently, in~\cite{liang-li} the authors propose 
ODIN, a simple method applied to DNNs that uses a softmax layer for classification and does not need the network to be retrained. The key idea is to use temperature scaling and input pre-processing, which consists on introducing small perturbations in the direction of the gradients for the input images.

In~\cite{lee-lee} they diverge from the other threshold-based methods by proposing a new training method. They add two loss terms that force the out-of-distribution samples to be less confident and improve the in-distribution samples respectively. In both these works, trained DNNs follow a typical softmax cross-entropy classification loss, where each dimension on the output embedding is assigned to measure the correlation with a specific class from that task. Other than previous work which focuses on networks trained with the cross-entropy, our work studies out-of-distribution for networks which are optimized for metric learning. These networks do not have the normalization problem which is introduced by the softmax layer, and are therefore expected to provide better estimates of out-of-distribution data.
One last work is still worth to mention in the context of DNNs. In~\cite{schultheiss-kading} the authors propose to discern between seen and unseen classes through the dimensions of certain layer activations which have extreme values. They achieve a good accuracy on ImageNet but only when the number of selected classes is very small.

\minisection{Open Set Recognition.} It shares with out-of-distribution detection the goal of discriminating samples from two different distributions. But it places the emphasis on how to apply it to improve the classifier capabilities, so that it can still perform well when the input may contain samples not belonging to any of those in the training set. One of the first works is~\cite{Scheirer13}, which formalized the problem as one of (open) risk minimization in the context of large margin classifiers, producing what they called a one-versus-set Support Vector Machine. More recently, a method to adapt deep neural networks to handle open set recognition has been proposed in~\cite{bendale-boult16}. The key idea is to replace the conventional softmax layer in a network by a so called openmax layer. It takes the $N$ activations (being $N$ the number of classes) of the penultimate layer of the network and estimates the probability for each training class, like in softmax, plus that of not being a sample of the training data. This later is done by fitting a Weilbull density function to the distance between the mean activation value for each class and those of the training samples. We see thus that distance between last layer activations or features play a key role. This is coincident with our method, only that features in their case are learned through a loss function similar to cross-entropy whereas we explicitly will learn a distance such that in-distribution samples cluster around one center per class and out-of-distribution samples are pushed away from all these centers.

\minisection{Metric Learning.} Several computer vision tasks such as retrieval, matching, verification, even multi-class classification, share the need of being able to measure the similarity between pairs of images. Deriving such a measure from data samples is known as metric learning~\cite{kulis2013metric}. Two often cited seminal works on this subject through neural networks are~\cite{chopra2005learning,Hadsell06}, where the Siamese architecture was proposed for this purpose. Differently from classification networks, the goal is to learn rather than a representation amenable for classification, one for measuring how similar two instances are in terms of the Euclidean distance. Another popular architecture is triplet networks~\cite{hoffer2015deep}. For both of them many authors have realized that mining the samples of the training set in order to find out \textit{difficult} or challenging pairs or triplets is important in order to converge faster or to better minima~\cite{Schroff15,song2016deep,Sohn16}. Like them, we have also resorted to a mining strategy in order to obtain good results in the task of out-of-distribution detection.


\section{Metric Learning for Out-of-Distribution}
Most recent works on out-of-distribution detection are based on supervisely trained neural networks which optimize the cross-entropy loss. In these cases the network output has a direct correspondence with the solution of the task, namely a probability for each class. However, the representation of the output vector is forced to always sum up to one. This means that when the network is shown an input which is not part of the training distribution, it will still give probabilities to the nearest classes so that they sum up to one. This phenomena has led to the known problem of neural networks being too overconfident about content that they have never seen~\cite{hendrycks-gimpel}.

Several works have focused on improving the accuracy of the confidence estimate of methods based on the cross entropy; adapting them in such a way that they would yield lower confidences for out-of-distribution~\cite{hendrycks-gimpel,liang-li,lee-lee}. We hypothesize that the problem of the overconfident network predictions is inherent to the used cross-entropy, and therefore propose to study another class of network objectives, namely those used for metric learning. In metric learning methods, we minimize an objective which encourages images with the same label to be close and images with different labels to be at least some margin apart in an embedding space. These networks do not apply a softmax layer, and therefore are not forced to divide images which are out-of-distribution over the known classes.

\subsection{Metric Learning}
For applications such as image retrieval, images are represented by an embedding in some feature space. Images can be ordered (or classified) according to the distance to other images in that embedding space. It has been shown that using metric learning methods to improve the embeddings could significantly improve their performance~\cite{guillaumin2009you}. The theory of metric learning was extended to deep neural networks by Chopra et al.~\cite{chopra2005learning}. They proposed to pass images through two parallel network branches which share the weights (also called a Siamese network). A loss considers both embeddings, and adapts the embedding in such a way that similar classes are close and dissimilar classes are far in that embedding space. 

Traditionally these networks have been trained with contrastive loss~\cite{Hadsell06}, which is formulated as:
\begin{equation}
L(x_1,x_2,y \, ; \, W) = \frac{1}{2}\left( {1 - y} \right)D_w^2  + \frac{1}{2}y\left( {\max \left( {0,m - D_w } \right)} \right)^2 \label{eq:contr},
\end{equation}
where $D_w  = ||{f_W \left( {x_1 } \right) - f_W \left( {x_2 } \right)}||_2$ is the distance between the embeddings of images $x_1$ and $x_2$ computed by network $f_W$ with weights $W$. The label $y=0$ indicates that the two images are from the same class, and $y=1$ is used for images from different classes. The loss therefore minimizes the distance between images of the same class, and increases the distance of images of different classes until this distance surpasses the margin $m$. Several other losses have been proposed for Siamese networks~\cite{hoffer2015deep,wang2014learning,Schroff15,wang2017deep,song2016deep} but in this paper we will evaluate results with the contrastive loss to provide a simple baseline on which to improve.

\subsection{Out-of-Distribution Mining (ODM)}
In the previous section, we considered that during training only examples of in-distribution data are provided. However, some methods consider the availability of some out-of-distribu\-tion data during training~\cite{lee-lee}. This is often a realistic assumption since it is relatively easy to obtain data from other datasets or create out-of-distribution examples, such as samples generated with Gaussian noise. However, it has to be noted that the out-of-distribution data is used unlabeled, and is of a different distribution from the out-of-distribution used at testing. The objective is to help the network be less confident about what it does not know. Therefore, noise or even unlabeled data can be used to strengthen the knowledge boundaries of the network.

We propose to adapt the contrastive loss to incorporate the out-of-distribution data:
\begin{equation}
L(x_1,x_2,y \, ; \, W) = \frac{1}{2}\left( {1 - y} \right)zD_w^2  + \frac{1}{2}yz\left( {\max \left( {0,m - D_w } \right)} \right)^2 \label{eq:contr2},
\end{equation}
where we have introduced a label $z$ which is zero when both images are from the out-of-distribution and one otherwise. This loss is similar to Eq.~\ref{eq:contr}, but with the difference that in case of a pair of images where one is an out-of-distribution image ($z=1$, $y=1$) they are encouraged to be at least $m$ distance apart. Note that we do not enforce the out-of-distribution images to be close, since when $z=0$ the pair does not contribute to the loss. It is important to make sure that there are no pairs of out-of-distribution samples so that they are not treated as a single new class and forced to be grouped into a single cluster.

In practice, we have not implemented a two-branches Siamese network but followed recent works~\cite{ustinova2016learning,liu2017rankiqa} which devise a more efficient approach to minimize losses traditionally computed with Siamese networks. The idea is to sample a minibatch of images which we forward through a single branch until the embedding layer. We then sample pairs from them in the loss layer and backpropagate the gradient. This allows the network to be defined with only one copy of the weights instead of having two branches with shared weights. At the same time, computing the pairs after the embedding also allows to use any subgroup of possible pairs among all the images from the minibatch. When computing the pairs we make sure that pairs of out-of-distribution samples are not used. As a result $z$ will never be $0$ and we can in practice directly apply Eq.~\ref{eq:contr} instead of Eq.~\ref{eq:contr2}.

\subsection{Anomaly and Novelty detection}
In this paper we distinguish between two categories of out-of-distribution data:

\indent\begin{minipage}{0.9\textwidth}
\minisection{Novelty:} samples that share some common space with the trained distribution, which are usually concepts or classes which the network could include when expanding its knowledge. If you train a network specialized in different dog breeds, an example would be a new dog breed that was not in the training set. Furthermore, if the classes are more complex, some novelty out-of-distribution could be new viewpoints or modifications of an existing learned class.

\minisection{Anomaly:} samples that are not related with the trained distribution. In this category we could include background images, Gaussian noise, or unrelated classes to the trained distribution (i.e. SVHN would be a meaningful anomaly for CIFAR-10). Since anomalies are further from the in-distribution than novelties these are expected to be easier to detect.
\end{minipage}
\smallskip

To further illustrate the difference between novelties and anomalies consider the following experiment. We train a LeNet on the classes 2, 6 and 7 from the MNIST dataset~\cite{lecun1998gradient} under the same setup for both cross-entropy~(CE) and contrastive~(ML) losses. We also train it with our proposed method which introduces out-of-distribution mining during training~(ODM). We use classes 0, 3, 4, and 8 as those seen out-of-distribution samples during training. Then, we visualize the embeddings for different out-of-distribution cases from closer to further resemblance to the train set~: 1) similar numbers 5, 9 and 1 as novelty, 2) SVHN~\cite{netzer2011reading} and CIFAR-10~\cite{krizhevsky2009learning} as anomalies with a meaning, and 3) the simpler Gaussian noise anomalies.

\begin{figure}
\centering
\includegraphics[width=.17\textwidth]{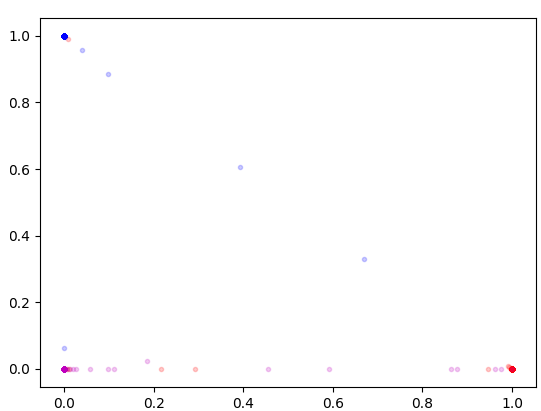}~
\includegraphics[width=.17\textwidth]{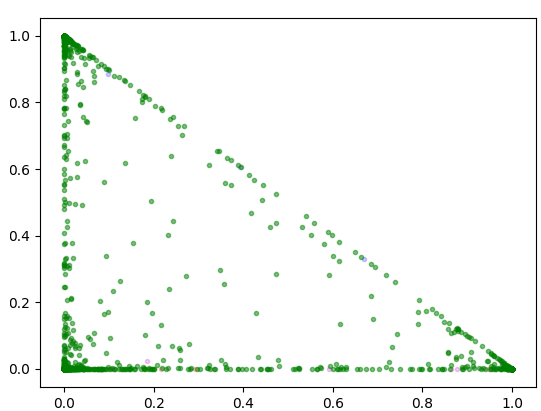}~
\includegraphics[width=.17\textwidth]{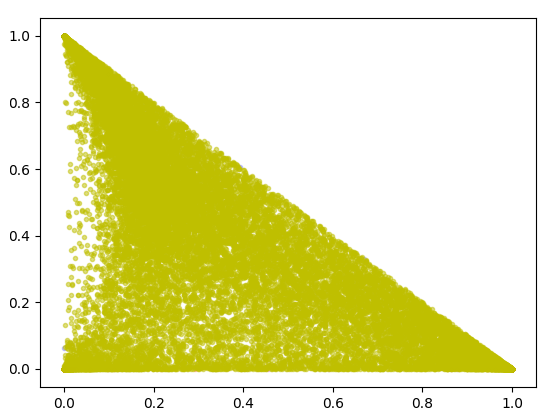}~
\includegraphics[width=.17\textwidth]{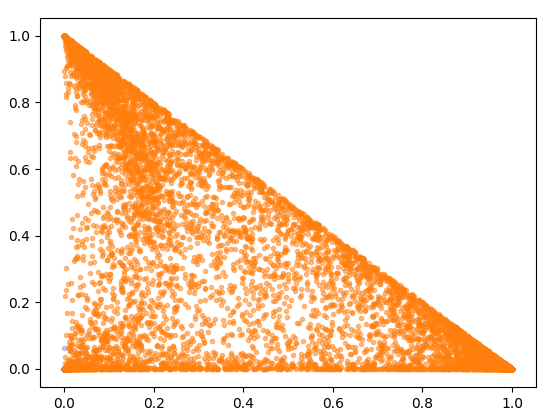}~
\includegraphics[width=.17\textwidth]{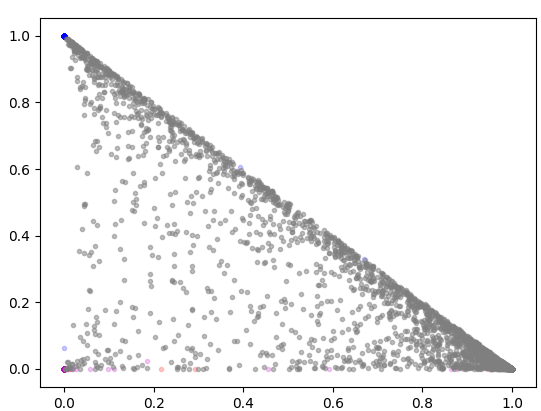}
\\
\includegraphics[width=.17\textwidth]{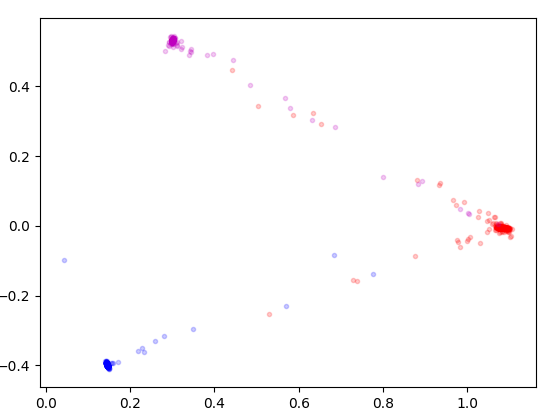}~
\includegraphics[width=.17\textwidth]{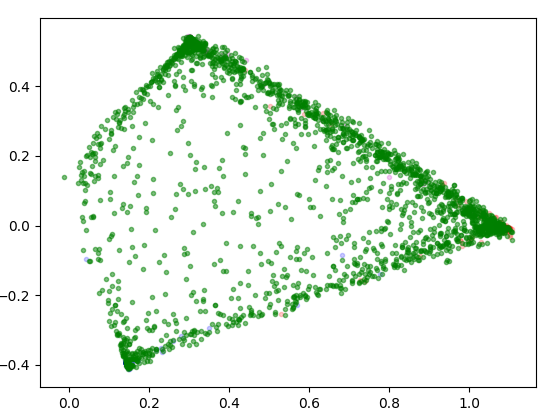}~
\includegraphics[width=.17\textwidth]{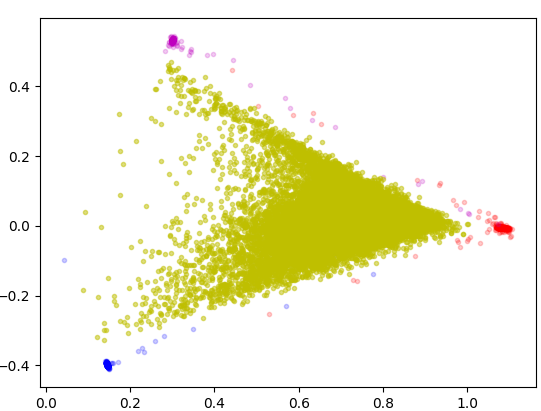}~
\includegraphics[width=.17\textwidth]{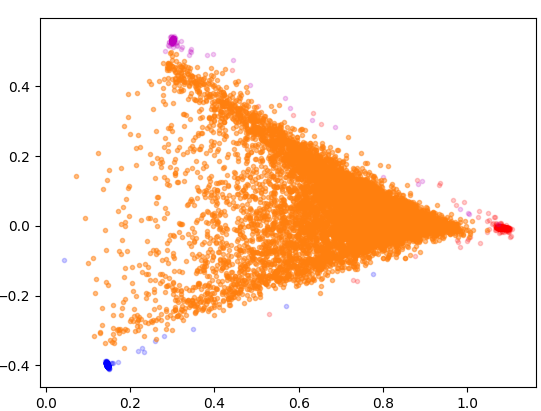}~
\includegraphics[width=.17\textwidth]{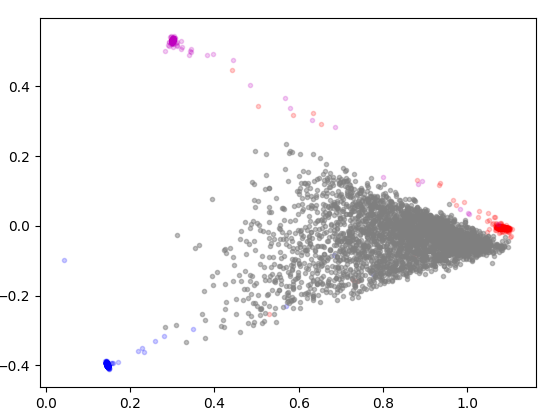}
\\
\includegraphics[width=.17\textwidth]{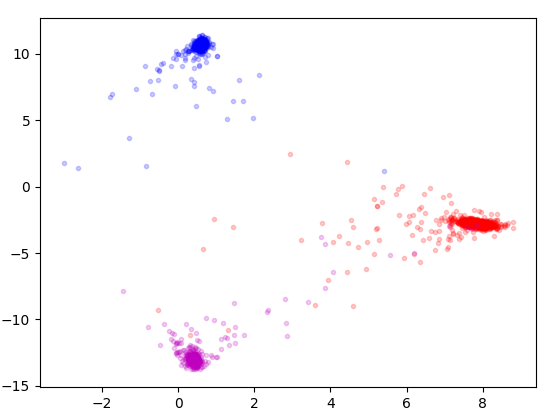}~
\includegraphics[width=.17\textwidth]{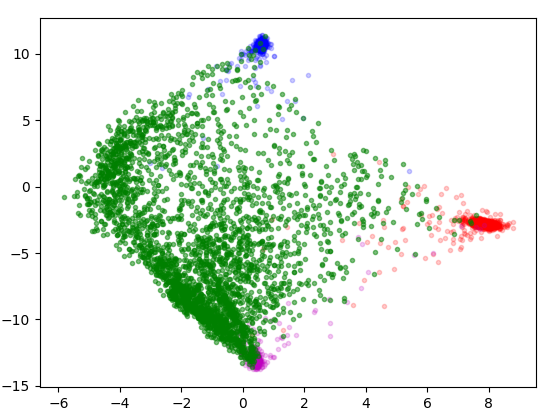}~
\includegraphics[width=.17\textwidth]{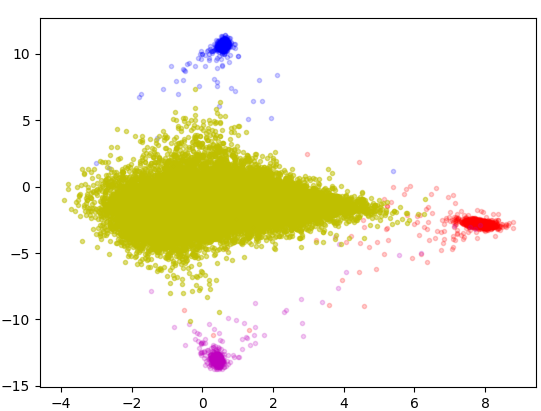}~
\includegraphics[width=.17\textwidth]{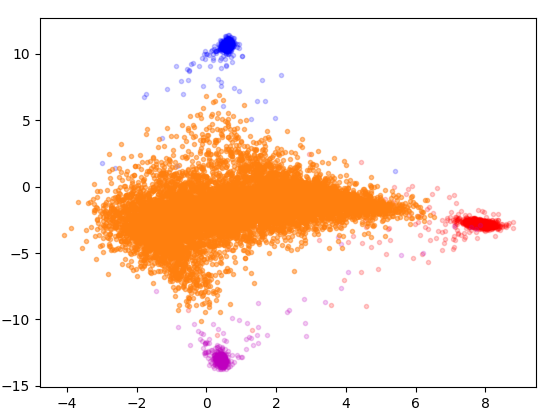}~
\includegraphics[width=.17\textwidth]{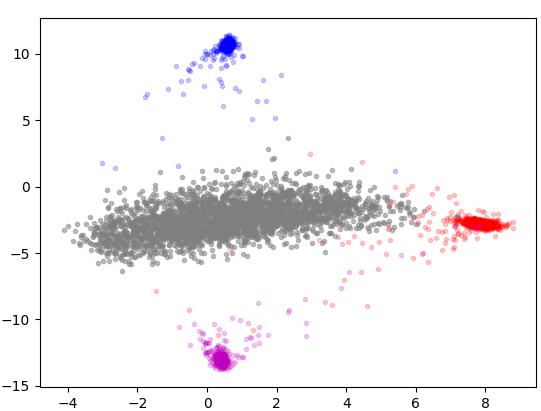}
\caption{Embedding spaces for CE, ML and ODM (rows respectively) being tested on in-dist 2, 6, 7 of MNIST (red, blue, purple), and out-dist 5, 9, 1 of MNIST (green), SVHN (yellow), CIFAR-10 (orange), and Gaussian noise (grey). Best viewed in color. }
\label{fig:motivation}
\end{figure}

In Figure~\ref{fig:motivation} we show the 3-dimensional output embedding spaces for CE, ML and ODM in rows 1, 2 and 3 respectively. As expected, the CE space is bounded inside the shown triangle, since the three dimensions of the output (the number of classes) have to always sum up to 1. For SVHN, CE correctly assigns low confidence for all classes. However, for CIFAR-10, Gaussian noise and Novelty it increasingly is more confident about the probability of an out-of-distribution image to be classified as an in-distribution one. In the case of ML, all anomalies seem to be more separated from the in-distributions for each class, and only the Novelty is still too close to the cluster centers. With the introduction of out-of-distribution samples during training, ODM shows how out-of-distribution images are kept away from the in-distribution, allowing the network to be confident about what it is capable of classifying and what not.  We provide quantitative performance results for this experiment in the Supplementary Material.

In conclusion, this experiment shows that there is a difference between novel and anoma\-ly out-of-distribution samples for both cross-entropy and metric learning approaches, stressing that those have to be approached differently. Furthermore, the overconfidence of the cross-entropy methods is more clear on novelty detection cases, and among the anomaly cases, the Gaussian noise seems to be the one with more overconfident cases. In those cases, a metric learning approach presents more benefits when doing out-of-distribution detection. It allows for the output embedding space to be more representative of the learned classes around the class centers, and naturally has the ability to give low scores to unseen data. Finally, when some out-of-distribution samples are shown during training, the network is more capable of adapting the embedding space to be more separable against anomaly data.

\section{Results}
To assess the performance of the proposed method, we first compare with existing state-of-the-art out-of-distribution detection methods on SVHN~\cite{netzer2011reading} and CIFAR-10~\cite{krizhevsky2009learning} datasets trained on VGGnet~\cite{szegedy2015going} and evaluated with the metrics provided in \cite{lee-lee}. Furthermore, as a more application-based benchmark, we propose to compare cross-entropy based strategies and metric learning strategies on the Tsinghua dataset~\cite{zhu2016traffic} of traffic signs. In this second set of experiments we use our own implementation of the metrics defined in~\cite{liang-li}. More about the metrics used can be found in the Supplementary Material.\footnote{Code available at: \url{https://mmasana.github.io/OoD_Mining}}

\subsection{Comparison with state-of-the-art} \label{sec:compareSOA}
We compare our method with two very recent state-of-the-art methods. One of them uses a confidence classifier and an adversarial generator (CC-AG)~\cite{lee-lee} and like ours uses out-of-distribution images during training. The second method is ODIN~\cite{liang-li} which does not consider out-of-distribution images during training. In~\cite{lee-lee} they compare CC-AG with ODIN~\cite{liang-li}, and show that they can perform much better in the novelty case but similar for the anomaly cases.

We train each SVHN and CIFAR-10 as the in-distribution datasets while using the other dataset as the seen out-distribution during training. We train on VGGnet, just like~\cite{lee-lee}, with a contrastive loss of margin 10 and a 25\% of (in-dist, out-dist) pairs every two batches. Following the experiments of~\cite{lee-lee}, we test the resulting networks on the in-distribution test set for classification, and TinyImageNet~\cite{deng2009imagenet}, LSUN~\cite{yu15lsun} and Gaussian noise for out-of-distribution detection. For evaluation we use the proposed metrics from their implementation, namely: true negative rate (TNR) when true positive rate (TPR) is at 95\%, detection accuracy, area under the receiver operating characteristic curve (AUROC) and both area under the precision-recall curve for in-distribution (AUPR-in) and out-distribution (AUPR-out).

Table~\ref{table:experiment1} shows the results. For SVHN as the in-distribution results are as expected, with ODIN having lower results due to not using any out-of-distribution during training, and both CC-AG and ODM having near perfect performance. In the case of CIFAR-10 being the in-distribution, the same pattern is repeated for the seen distribution from SVHN. However, for the unseen out-distributions, CC-AG achieves the lower performance on both TinyImageNet and LSUN datasets, and ODIN the lower for Gaussian noise. Although not always achieving the best performance, ODM is able to compete with the best cases, and is never the worse performer. Gaussian noise seems to be the most difficult case on CIFAR-10, which is a more complex dataset than SVHN. For ODIN, as it is only based on cross-entropy, it becomes to overconfident. In the case of CC-AG and ODM, the low results might be related to Gaussian noise being too different from the out-distribution seen during training.

Finally, it is important to note that metric learning has a lower classification accuracy of the in-distribution. This has already been observed in~\cite{horiguchi-ikami}, where features learned by classification networks with typical softmax layers are compared with metric learning based features, with regard to several benchmark datasets. For good classification results our metric learning network should be combined with those of a network trained with cross-entropy. One could also consider a network with two heads, where after some initial shared layers a cross-entropy branch and a metric learning branch are trained in a multi-task setting.

\begin{table}
\begin{center}
\caption{Comparison with the state-of-the-art. All metrics show the methods as ODIN/CC-AG/ODM, red indicates worst performance, bold indicates best, * for seen distribution.}
\label{table:experiment1}
\resizebox{\textwidth}{!}{
\begin{tabular}{ccccccc}
\hline
\noalign{\smallskip}
In-dist        & \multirow{ 2}{*}{Out-dist} & TNR at   & Detection & \multirow{ 2}{*}{AUROC} & \multirow{ 2}{*}{AUPR-in} & \multirow{ 2}{*}{AUPR-out} \\
classification &                            & 95\% TPR & Accuracy     &  &  & \\
\hline
               & CIFAR-10*       & \textcolor{red}{47.4}/\textbf{99.9}/99.8   & \textcolor{red}{78.6}/\textbf{99.9}/99.8   & \textcolor{red}{62.6}/\textbf{99.9}/99.5   & \textcolor{red}{71.6}/\textbf{99.9}/99.7   & \textcolor{red}{91.2}/99.4/\textbf{99.9}  \\
SVHN           & Tiny & \textcolor{red}{49.0}/\textbf{100.0}/99.0  & \textcolor{red}{79.6}/\textbf{100.0}/99.1  & \textcolor{red}{64.6}/\textbf{100.0}/99.0  & \textcolor{red}{72.7}/\textbf{100.0}/96.5  & \textcolor{red}{91.6}/99.4/\textbf{99.8}  \\
93.8/\textbf{94.2}/\textcolor{red}{68.7} & LSUN         & \textcolor{red}{46.3}/\textbf{100.0}/99.4  & \textcolor{red}{78.2}/\textbf{100.0}/99.5  & \textcolor{red}{61.8}/\textbf{100.0}/99.3  & \textcolor{red}{71.1}/\textbf{100.0}/97.8  & \textcolor{red}{90.8}/99.4/\textbf{99.8}  \\
               & Gaussian     & \textcolor{red}{56.1}/\textbf{100.0}/\textbf{100.0} & \textcolor{red}{83.4}/\textbf{100.0}/\textbf{100.0} & \textcolor{red}{72.0}/\textbf{100.0}/\textbf{100.0} & \textcolor{red}{77.2}/\textbf{100.0}/\textbf{100.0} & \textcolor{red}{92.8}/99.4/\textbf{100.0} \\
\hline
               & SVHN*           & \textcolor{red}{13.7}/\textbf{99.8}/\textbf{99.8} & \textcolor{red}{66.6}/\textbf{99.8}/99.7 & \textcolor{red}{46.6}/\textbf{99.9}/\textbf{99.9} & \textcolor{red}{61.4}/\textbf{99.9}/\textbf{99.9} & \textcolor{red}{73.5}/99.8/\textbf{100.0} \\
CIFAR-10       & Tiny & 13.6/\textcolor{red}{10.1}/\textbf{17.1} & 62.6/\textcolor{red}{58.9}/\textbf{66.9} & 39.6/\textcolor{red}{31.8}/\textbf{66.2} & 58.3/\textcolor{red}{55.3}/\textbf{60.3} & \textbf{71.0}/\textcolor{red}{66.1}/68.2 \\
80.1/\textbf{80.6}/\textcolor{red}{54.0} & LSUN         & 14.0/\textcolor{red}{10.8}/\textbf{19.6} & 63.2/\textcolor{red}{60.2}/\textbf{70.9} & 40.7/\textcolor{red}{34.8}/\textbf{68.4} & 58.7/\textcolor{red}{56.4}/\textbf{59.5} & \textbf{71.5}/\textcolor{red}{68.0}/70.7 \\
               & Gaussian     & \textcolor{red}{2.8}/\textbf{3.5}/3.0   & \textcolor{red}{50.0}/\textcolor{red}{50.0}/\textbf{64.2} & \textcolor{red}{10.2}/14.1/\textbf{49.8} & \textcolor{red}{48.1}/49.4/\textbf{64.1} & \textcolor{red}{39.9}/\textbf{47.0}/46.7 \\
\hline
\end{tabular}}
\end{center}
\end{table}

\subsection{Tsinghua traffic sign dataset} \label{sec:tsinghua_results}
\begin{figure}
\centering
\begin{minipage}{.3\linewidth}
 \centering
    \includegraphics[width=\textwidth]{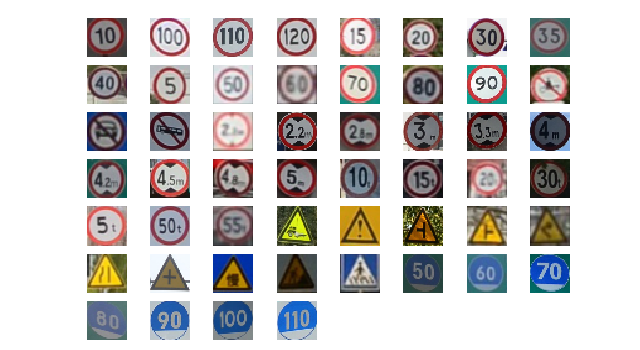}
\end{minipage}
\begin{minipage}{.3\linewidth}
 \centering
    \includegraphics[width=\textwidth]{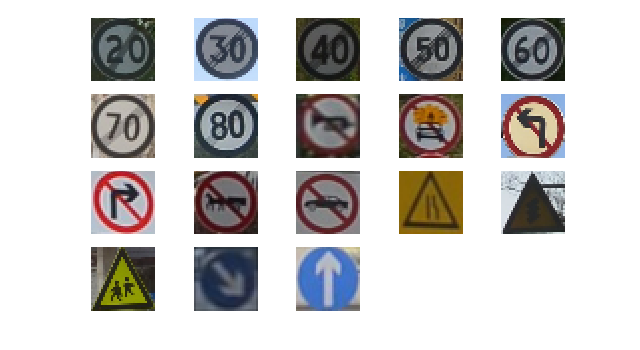}
\end{minipage}
\begin{minipage}{.3\linewidth}
 \centering
    \includegraphics[width=\textwidth]{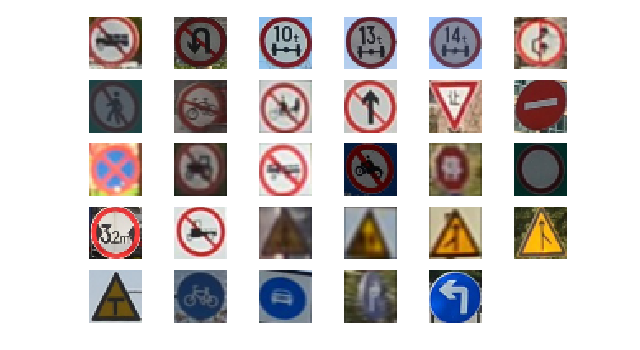}
\end{minipage}
    \caption{In-distribution (left), seen (middle) and unseen(right) out-of-distribution partition classes from the proposed Tsinghua split.}
    \label{fig:groups_traffic_signs}
\end{figure}

We evaluate our method on a real application, \textit{i.e.} traffic sign recognition in the presence of unseen traffic signs (novelty) and not-a-traffic-sign detection (anomaly). We compare our proposed method ODM against ODIN~\cite{liang-li}, as a cross-entropy based method, on the Tsinghua dataset~\cite{zhu2016traffic}. We divide traffic sign classes into three disjoint partitions~: the in-distribution classes, seen out-of-distribution images used for training, and unseen out-of-distribution images used for testing on out-of-distribution detection. Since Tsinghua contains some very similar traffic sign classes which would rarely be learned without each other (i.e. all speed limits, all turning arrows, ...), we group those that are too similar in order to build a more reasonable and natural split than just a random one (See Supplementary Material for more on the usual random splits). For the same reason, we also discard classes with less than 10 images as they introduce errors. Therefore, we generate a random split which applies by the mentioned restrictions (see Fig.~\ref{fig:groups_traffic_signs}), by taking a 50-20-30\% split of the classes for the in-distribution, seen out-distribution and unseen out-distribution respectively.

Regarding anomalies, we consider Gaussian noise, but also background patches from the same Tsinghua dataset images. Those patches are generated randomly from the central area of the original full frames to avoid an unbalanced ratio of ground and sky images, which can be semantically richer and more challenging. In a real traffic sign detector application, where detected possible traffic signs are fed to a classifier, this kind of anomalies are more realistic and account for possible detection errors more than Gaussian noise. The global performance of the system can be improved by avoiding that those anomalies reach the classifier and produce an overconfident error.

For this experiment, we learn a 32-dimensional embedding space, training a WRN-28-10 model~\cite{zagoruyko2016wide} with an Adam optimizer at learning rate 0.0001 for 10,000 steps. The same training parameters are used for ODIN since they provided the best combination on the validation set. Table~\ref{table:tsinghua_restricted} shows the results of the comparison between ODIN, ML and ODM for both seen novelty and anomaly cases. Note that our implementation of the Detection Error metric is fixed to use the FPR at a TPR of 95\%, making a value of 2.50 the one of a perfect detector (see Supplementary Material).

In terms of in-distribution classification accuracy, both methods are equivalent. However, the comparison of plain metric learning (Ours-ML) with ODIN shows that learning an embedding can be more suitable for out-of-distribution detection of both novelty and anomalies. Introducing out-distribution samples during training slightly improves all cases. Using anomalies as seen out-of-distribution during training helps the detection of the same kind of anomaly as expected since anomalies will be forced to be further away from the in-distribution in the embedding space. However, in some cases, it can damage the detection of novelty, which would not be guaranteed to be pushed away from the learned classes.

\begin{table}
\begin{center}
\caption{Comparison between ODIN and our proposed learning strategies on a WRN-28-10 architecture, when using novelty, anomaly (background patches and Gaussian noise) as seen out-of-distribution data as well as not seen out-of-distribution.}
\label{table:tsinghua_restricted}
\resizebox{0.95\textwidth}{!}{
\begin{tabular}{ccccccccc}
\hline
\noalign{\smallskip}
\multirow{ 2}{*}{Method}  & In-dist  & \multirow{ 2}{*}{Out-dist} & FPR at    & Detection    & \multirow{ 2}{*}{AUROC} & \multirow{ 2}{*}{AUPR-in} & \multirow{ 2}{*}{AUPR-out} \\
                          & accuracy &                            & 95\% TPR  & error &  &  & \\
\hline
\multirow{ 3}{*}{ODIN}    & \multirow{ 3}{*}{98.29} & Tsinghua (unseen)   &  8.74 &  6.87 & 97.82 & 96.19 & 98.92 \\
                          &                         & Background (unseen) & 22.42 & 13.71 & 96.43 & 92.13 & 98.48 \\
                          &                         & Noise (unseen)      &  0.23 &  2.61 & 98.59 & 98.40 & 98.76 \\
\hline
\multirow{ 3}{*}{Ours - ML}    & \multirow{ 3}{*}{98.93} & Tsinghua (unseen)   & 5.23 & 5.11 & 98.77 & 97.38 & 99.45 \\
                          &                         & Background (unseen) & 0.25 & 2.62 & 99.35 & 99.03 & 99.64 \\
                          &                         & Noise (unseen)      & 0.07 & 2.53 & 99.51 & 99.25 & 99.72 \\
\hline
\multirow{ 3}{*}{Ours - ODM}    & \multirow{ 3}{*}{98.96} & \textbf{Tsinghua} (seen)   & 4.38 & 4.70 & 99.01 & 98.01 & 99.63 \\
                          &                         & Background (unseen)        & 0.17 & 2.60 & 99.28 & 98.81 & 99.67 \\
                          &                         & Noise (unseen)             & 0.00 & 2.51 & 99.69 & 99.51 & 99.73 \\
\hline
\multirow{ 3}{*}{Ours - ODM}    & \multirow{ 3}{*}{98.57} & Tsinghua (unseen)          & 8.65 & 6.82 & 97.84 & 94.40 & 98.57 \\
                          &                         & \textbf{Background} (seen) & 0.01 & 2.50 & 99.99 & 99.94 & 99.99 \\
                          &                         & Noise (unseen)             & 0.00 & 2.50 & 100.00 & 99.97 & 99.99 \\
\hline
\multirow{ 3}{*}{Ours - ODM}    & \multirow{ 3}{*}{99.00} & Tsinghua (unseen)          & 5.72 & 5.36 & 98.50  & 97.09 & 99.30 \\
                          &                         & Background (unseen)        & 1.51 & 3.25 & 98.53  & 97.97 & 99.20 \\
                          &                         & \textbf{Noise} (seen)      & 0.00 & 2.50 & 100.00 & 99.93 & 99.99 \\
\hline
\end{tabular}}
\end{center}
\end{table}

\section{Conclusions}
In this paper, we propose a metric learning approach to improve out-of-distribution detection which performs comparable or better than the state-of-the-art. We show that metric learning provides a better output embedding space to detect data outside the learned distribution than cross-entropy softmax based models. This opens an opportunity to further research on how this embedding space should be learned, with restrictions that could further improve the field. The presented results suggest that out-of-distribution data might not all be seen as a single type of anomaly, but instead a continuous representation between novelty and anomaly data. In that spectrum, anomaly detection is the easier task, giving more focus at the difficulty of novelty detection. Finally, we also propose a new benchmark for out-of-distribution detection on the Tsinghua dataset, as a more realistic scenario for novelty detection. 

\section*{Acknowledgements}
Marc Masana acknowledges 2018-FI\_B1-00198 grant of \textit{Generalitat de Catalunya}. Idoia Ruiz, Joan Serrat and Antonio Lopez want to acknowledge the Spanish project TIN2017-88709-R (Ministerio de Ciencia, Innovaci\'{o}n y Universidades). This work is supported by the EU Project CybSpeed MSCA-RISE-2017-777720. We acknowledge the project TIN2016-79717-R and the CHISTERA project M2CR (PCIN-2015-251) of the Spanish Government. We also acknowledge the CERCA Programme of Generalitat de Catalunya and its ACCIO agency. Finally, we acknowledge the generous support of the NVIDIA GPU donation program.

\bibliography{our_submission}

\begin{thebibliography}{35}
\providecommand{\natexlab}[1]{#1}
\providecommand{\url}[1]{\texttt{#1}}
\expandafter\ifx\csname urlstyle\endcsname\relax
  \providecommand{\doi}[1]{doi: #1}\else
  \providecommand{\doi}{doi: \begingroup \urlstyle{rm}\Url}\fi

\bibitem[Bendale and Boult(2016)]{bendale-boult16}
A.~Bendale and T.~E. Boult.
\newblock Towards open set deep networks.
\newblock In \emph{IEEE Conference on Computer Vision and Pattern Recognition
  (CVPR)}, pages 1563--1572, 2016.

\bibitem[Bodesheim et~al.(2013)Bodesheim, Freytag, Rodner, Kemmler, and
  Denzler]{Bodesheim13}
Paul Bodesheim, Alexander Freytag, Erik Rodner, Michael Kemmler, and Joachim
  Denzler.
\newblock Kernel null space methods for novelty detection.
\newblock In \emph{IEEE Conference on Computer Vision and Pattern Recognition
  (CVPR)}, pages 3374--3381, 2013.

\bibitem[Chen and Liu(2016)]{chen2016lifelong}
Zhiyuan Chen and Bing Liu.
\newblock Lifelong machine learning.
\newblock \emph{Synthesis Lectures on Artificial Intelligence and Machine
  Learning}, 10\penalty0 (3):\penalty0 1--145, 2016.

\bibitem[Chopra et~al.(2005)Chopra, Hadsell, and LeCun]{chopra2005learning}
Sumit Chopra, Raia Hadsell, and Yann LeCun.
\newblock Learning a similarity metric discriminatively, with application to
  face verification.
\newblock In \emph{IEEE Conference on Computer Vision and Pattern Recognition
  (CVPR)}, pages 539--546, 2005.

\bibitem[Davis and Goadrich(2006)]{davis2006relationship}
Jesse Davis and Mark Goadrich.
\newblock The relationship between precision-recall and roc curves.
\newblock In \emph{Proceedings of the 23rd international conference on Machine
  learning}, pages 233--240. ACM, 2006.

\bibitem[Deng et~al.(2009)Deng, Dong, Socher, Li, Li, and
  Fei-Fei]{deng2009imagenet}
Jia Deng, Wei Dong, Richard Socher, Li-Jia Li, Kai Li, and Li~Fei-Fei.
\newblock Imagenet: A large-scale hierarchical image database.
\newblock In \emph{IEEE Conference on Computer Vision and Pattern Recognition
  (CVPR)}, pages 248--255, 2009.

\bibitem[Geifman and El{-}Yaniv(2017)]{geifman-yaniv}
Yonatan Geifman and Ran El{-}Yaniv.
\newblock Selective classification for deep neural networks.
\newblock In \emph{Advances in Neural Information Processing Systems (NIPS)},
  pages 4885--4894, 2017.

\bibitem[Guillaumin et~al.(2009)Guillaumin, Verbeek, and
  Schmid]{guillaumin2009you}
Matthieu Guillaumin, Jakob Verbeek, and Cordelia Schmid.
\newblock Is that you? metric learning approaches for face identification.
\newblock In \emph{Computer Vision, 2009 IEEE 12th international conference
  on}, pages 498--505. IEEE, 2009.

\bibitem[Hadsell et~al.(2006)Hadsell, Chopra, and LeCun]{Hadsell06}
Raia Hadsell, Sumit Chopra, and Yann LeCun.
\newblock Dimensionality reduction by learning an invariant mapping.
\newblock In \emph{IEEE Conference on Computer Vision and Pattern Recognition
  (CVPR)}, pages 1735--1742, 2006.

\bibitem[Hendrycks and Gimpel(2017)]{hendrycks-gimpel}
Dan Hendrycks and Kevin Gimpel.
\newblock A baseline for detecting misclassified and out-of-distribution
  examples in neural networks.
\newblock In \emph{Int. Conference on Learning Representations (ICLR)}, 2017.

\bibitem[Hoffer and Ailon(2015)]{hoffer2015deep}
Elad Hoffer and Nir Ailon.
\newblock Deep metric learning using triplet network.
\newblock In \emph{International Workshop on Similarity-Based Pattern
  Recognition}, pages 84--92. Springer, 2015.

\bibitem[Horiguchi et~al.(2017)Horiguchi, Ikami, and Aizawa]{horiguchi-ikami}
Shota Horiguchi, Daiki Ikami, and Kiyoharu Aizawa.
\newblock Significance of softmax-based features in comparison to distance
  metric learning-based features.
\newblock \emph{CoRR}, abs/1712.10151, 2017.

\bibitem[Kirkpatrick et~al.(2017)Kirkpatrick, Pascanu, Rabinowitz, Veness,
  Desjardins, Rusu, Milan, Quan, Ramalho, Grabska-Barwinska,
  et~al.]{kirkpatrick2017overcoming}
James Kirkpatrick, Razvan Pascanu, Neil Rabinowitz, Joel Veness, Guillaume
  Desjardins, Andrei~A Rusu, Kieran Milan, John Quan, Tiago Ramalho, Agnieszka
  Grabska-Barwinska, et~al.
\newblock Overcoming catastrophic forgetting in neural networks.
\newblock \emph{Proceedings of the national academy of sciences}, page
  201611835, 2017.

\bibitem[Krizhevsky and Hinton(2009)]{krizhevsky2009learning}
Alex Krizhevsky and Geoffrey Hinton.
\newblock Learning multiple layers of features from tiny images.
\newblock 2009.

\bibitem[Kulis(2013)]{kulis2013metric}
Brian Kulis.
\newblock Metric learning: A survey.
\newblock \emph{Foundations and Trends in Machine Learning}, 5\penalty0
  (4):\penalty0 287--364, 2013.

\bibitem[LeCun et~al.(1998)LeCun, Bottou, Bengio, and
  Haffner]{lecun1998gradient}
Yann LeCun, L{\'e}on Bottou, Yoshua Bengio, and Patrick Haffner.
\newblock Gradient-based learning applied to document recognition.
\newblock \emph{Proceedings of the IEEE}, 86\penalty0 (11):\penalty0
  2278--2324, 1998.

\bibitem[Lee et~al.(2018)Lee, Lee, Lee, and Shin]{lee-lee}
Kimin Lee, Honglak Lee, Kibok Lee, and Jinwoo Shin.
\newblock Training confidence-calibrated classifiers for detecting
  out-of-distribution samples.
\newblock In \emph{Int. Conference on Learning Representations (ICLR)}, 2018.

\bibitem[Liang et~al.(2018)Liang, Li, and Srikant]{liang-li}
Shiyu Liang, Yixuan Li, and R.~Srikant.
\newblock Enhancing the reliability of out-of-distribution image detection in
  neural networks.
\newblock In \emph{Int. Conference on Learning Representations (ICLR)}, 2018.

\bibitem[Liu et~al.(2017)Liu, van~de Weijer, and Bagdanov]{liu2017rankiqa}
Xialei Liu, Joost van~de Weijer, and Andrew~D Bagdanov.
\newblock Rankiqa: Learning from rankings for no-reference image quality
  assessment.
\newblock In \emph{International Conference on Computer Vision (ICCV)}, 2017.

\bibitem[Liu et~al.(2018)Liu, Masana, Herranz, Van~de Weijer, Lopez, and
  Bagdanov]{liu2018rotate}
Xialei Liu, Marc Masana, Luis Herranz, Joost Van~de Weijer, Antonio~M Lopez,
  and Andrew~D Bagdanov.
\newblock Rotate your networks: Better weight consolidation and less
  catastrophic forgetting.
\newblock In \emph{Proceedings International Conference on Pattern Recognition
  (ICPR)}, 2018.

\bibitem[Manning and Sch{\"u}tze(1999)]{manning1999foundations}
Christopher~D Manning and Hinrich Sch{\"u}tze.
\newblock \emph{Foundations of statistical natural language processing}.
\newblock MIT press, 1999.

\bibitem[Netzer et~al.(2011)Netzer, Wang, Coates, Bissacco, Wu, and
  Ng]{netzer2011reading}
Yuval Netzer, Tao Wang, Adam Coates, Alessandro Bissacco, Bo~Wu, and Andrew~Y
  Ng.
\newblock Reading digits in natural images with unsupervised feature learning.
\newblock In \emph{NIPS {W}orkshop on deep learning and unsupervised feature
  learning}, 2011.

\bibitem[Pimentel et~al.(2014)Pimentel, Clifton, Clifton, and
  Tarassenko]{pimentel-clifton}
Marco A.~F. Pimentel, David~A. Clifton, Lei~A. Clifton, and Lionel Tarassenko.
\newblock A review of novelty detection.
\newblock \emph{Signal Processing}, 99:\penalty0 215--249, 2014.

\bibitem[Scheirer et~al.(2013)Scheirer, de~Rezende~Rocha, Sapkota, and
  Boult]{Scheirer13}
W.~J. Scheirer, A.~de~Rezende~Rocha, A.~Sapkota, and T.~E. Boult.
\newblock Toward open set recognition.
\newblock \emph{IEEE Transactions on Pattern Analysis and Machine
  Intelligence}, 35\penalty0 (7):\penalty0 1757--1772, July 2013.

\bibitem[Schroff et~al.(2015)Schroff, Kalenichenko, and Philbin]{Schroff15}
Florian Schroff, Dmitry Kalenichenko, and James Philbin.
\newblock Facenet: A unified embedding for face recognition and clustering.
\newblock In \emph{IEEE Conference on Computer Vision and Pattern Recognition
  (CVPR)}, pages 815--823, 2015.

\bibitem[Schultheiss et~al.(2017)Schultheiss, K{\"a}ding, Freytag, and
  Denzler]{schultheiss-kading}
Alexander Schultheiss, Christoph K{\"a}ding, Alexander Freytag, and Joachim
  Denzler.
\newblock Finding the unknown: Novelty detection with extreme value signatures
  of deep neural activations.
\newblock In Volker Roth and Thomas Vetter, editors, \emph{German Conference on
  Pattern Recognition (GCPR)}, pages 226--238. Springer, 2017.

\bibitem[Sohn(2016)]{Sohn16}
Kihyuk Sohn.
\newblock Improved deep metric learning with multi-class n-pair loss objective.
\newblock In \emph{Advances in Neural Information Processing Systems (NIPS)},
  2016.

\bibitem[Song et~al.(2016)Song, Xiang, Jegelka, and Savarese]{song2016deep}
Hyun~Oh Song, Yu~Xiang, Stefanie Jegelka, and Silvio Savarese.
\newblock Deep metric learning via lifted structured feature embedding.
\newblock In \emph{IEEE Conference on Computer Vision and Pattern Recognition
  (CVPR)}, pages 4004--4012, 2016.

\bibitem[Szegedy et~al.(2015)Szegedy, Liu, Jia, Sermanet, Reed, Anguelov,
  Erhan, Vanhoucke, Rabinovich, et~al.]{szegedy2015going}
Christian Szegedy, Wei Liu, Yangqing Jia, Pierre Sermanet, Scott Reed, Dragomir
  Anguelov, Dumitru Erhan, Vincent Vanhoucke, Andrew Rabinovich, et~al.
\newblock Going deeper with convolutions.
\newblock In \emph{IEEE Conference on Computer Vision and Pattern Recognition
  (CVPR)}, 2015.

\bibitem[Ustinova and Lempitsky(2016)]{ustinova2016learning}
Evgeniya Ustinova and Victor Lempitsky.
\newblock Learning deep embeddings with histogram loss.
\newblock In \emph{Advances in Neural Information Processing Systems (NIPS)},
  pages 4170--4178, 2016.

\bibitem[Wang et~al.(2017)Wang, Zhou, Wen, Liu, and Lin]{wang2017deep}
Jian Wang, Feng Zhou, Shilei Wen, Xiao Liu, and Yuanqing Lin.
\newblock Deep metric learning with angular loss.
\newblock In \emph{International Conference on Computer Vision (ICCV)}, 2017.

\bibitem[Wang et~al.(2014)Wang, Leung, Rosenberg, Wang, Philbin, Chen, Wu,
  et~al.]{wang2014learning}
Jiang Wang, Thomas Leung, Chuck Rosenberg, Jinbin Wang, James Philbin, Bo~Chen,
  Ying Wu, et~al.
\newblock Learning fine-grained image similarity with deep ranking.
\newblock In \emph{IEEE Conference on Computer Vision and Pattern Recognition
  (CVPR)}, 2014.

\bibitem[Yu et~al.(2015)Yu, Zhang, Song, Seff, and Xiao]{yu15lsun}
Fisher Yu, Yinda Zhang, Shuran Song, Ari Seff, and Jianxiong Xiao.
\newblock Lsun: Construction of a large-scale image dataset using deep learning
  with humans in the loop.
\newblock \emph{arXiv preprint arXiv:1506.03365}, 2015.

\bibitem[Zagoruyko and Komodakis(2016)]{zagoruyko2016wide}
Sergey Zagoruyko and Nikos Komodakis.
\newblock Wide residual networks.
\newblock In \emph{British Machine and Vision Conference (BMVC)}, 2016.

\bibitem[Zhu et~al.(2016)Zhu, Liang, Zhang, Huang, Li, and Hu]{zhu2016traffic}
Zhe Zhu, Dun Liang, Songhai Zhang, Xiaolei Huang, Baoli Li, and Shimin Hu.
\newblock Traffic-sign detection and classification in the wild.
\newblock In \emph{IEEE Conference on Computer Vision and Pattern Recognition
  (CVPR)}, pages 2110--2118, 2016.

\end{thebibliography}

\newpage
\section*{\\ \huge \textbf{Supplementary Material} \\ \Large \textbf{Metric Learning for Novelty and Anomaly Detection}}

\noindent\makebox[\linewidth]{\rule{\textwidth}{0.5pt}} 

\renewcommand{\thesection}{\Alph{section}}
\setcounter{section}{0}

\section{Out-of-Distribution detection metrics}
In out-of-distribution detection, comparing different detector approaches cannot be done by measuring only accuracy. The question we want to answer is if a given test sample is from a different distribution than that of the training data. The detector will be using some information from the classifier or embedding space, but the prediction is whether that processed sample is part of the in-distribution or the out-distribution. To measure that, we adopt the metrics proposed in~\cite{liang-li}:
\begin{itemize}
    \item \textbf{FPR at 95\% TPR} is the corresponding False Positive Rate (FPR=FP/(FP+TN)) when the True Positive Rate (TPR=TP/(TP+FN)) is at 95\%. It can be interpreted as the misclassification probability of a negative (out-distribution) sample to be predicted as a positive (in-distribution) sample.
    \item \textbf{Detection Error} measures the probability of misclassifying a sample when the TPR is at 95\%. Assuming that a sample has equal probability of being positive or negative in the test, it is defined as $0.5(1-\textrm{TPR}) + 0.5\textrm{FPR}$.
\end{itemize}
where TP, FP, TN, FN correspond to true positives, false positives, true negatives and false negatives respectively. Those two metrics were also changed to \textbf{TNR at 95\% TPR} and \textbf{Detection Accuracy} in~\cite{lee-lee}, which can be calculated by doing $1-x$ from the two metrics above explained respectively. We use the latter metrics only when comparing to other state-of-the-art methods. This is also done because the implementation in both~\cite{liang-li,lee-lee} allows for using a TPR which is not at 95\% in some cases, meaning that the Detection Error can go below $2.5$ since TPR is not fixed to $0.95$.

In order to avoid the biases between the likelihood of an in-distribution sample to being more frequent than an out-distribution one, we need threshold independent metrics that measure the trade-off between false negatives and false positives. We adopt the following performance metrics proposed in~\cite{hendrycks-gimpel}:
\begin{itemize}
    \item \textbf{AUROC} is the Area Under the Receiver Operating Characteristic proposed in~\cite{davis2006relationship}. It measures the relation between between TPR and FPR interpreted as the probability of a positive sample being assigned a higher score than a negative sample.
    \item \textbf{AUPR} is the Area Under the Precision-Recall curve proposed in~\cite{manning1999foundations}. It measures the relationship between precision (TP/(TP+FP)) and recall (TP/(TP+FN)) and is more robust when positive and negative classes have different base rates. For this metric we provide both AUPR-in and AUPR-out when treating in-distribution and out-distribution samples as positive, respectively.
\end{itemize}
\section{Quantitative results of the MNIST experiment}
In this section we present the quantitative results of the comparison on the MNIST dataset. In this case we allowed a 5-dimensional embedding space for ML so the representation is rich enough to make the discrimination between in-dist and out-dist. For CE, as it is fixed to the number of classes, the embedding space is 3-dimensional. In Table~\ref{table:motivation} we see that ML performs a better than CE on all cases. ODM almost solves the novelty problem while keeping a similar performance on anomalies as ML. It is noticeable that CE struggles a bit more with Gaussian noise than the other anomalies. In this case, CE still produces highly confident predictions for some of the noise images. 
\begin{table}
\begin{center}
\caption{Quantitative comparison between cross-entropy and metric learning based methods training on LeNet for MNIST -- 2, 6, 7 (In-dist), 0, 3, 4 and 8 (Seen Out-dist) and 5, 9, 1 (Unseen Out-dist Novelty).}
\label{table:motivation}
\resizebox{0.95\textwidth}{!}{
\begin{tabular}{cccccccc}
\hline
\multirow{ 2}{*}{Method} & In-dist  & \multirow{ 2}{*}{Out-dist} & FPR at   & Detection & \multirow{ 2}{*}{AUROC} & \multirow{ 2}{*}{AUPR-in} & \multirow{ 2}{*}{AUPR-out} \\
                         & accuracy &                            & 95\% TPR & Error     &  &  & \\
\hline
\multirow{ 4}{*}{CE} & \multirow{ 4}{*}{99.70} & Novelty        & 33.76 & 19.38 & 92.33 & 92.73 & 92.29 \\
                     &                         & Gaussian noise &  0.70 &  2.85 & 98.85 & 99.21 & 98.14 \\
                     &                         & SVHN           &  0.23 &  2.60 & 99.48 & 98.64 & 99.91 \\
                     &                         & CIFAR-10       &  2.86 &  3.93 & 98.96 & 98.02 & 99.57 \\
\hline
\multirow{ 4}{*}{Ours - ML} & \multirow{ 4}{*}{99.54} & Novelty        & 21.05 & 13.03 & 94.48 & 94.02 & 94.46 \\
                     &                         & Gaussian noise &  0.00 & 1.95 & 98.54 & 99.21 & 95.15 \\
                     &                         & SVHN           &  0.00 & 1.74 & 98.88 & 98.76 & 99.61 \\
                     &                         & CIFAR-10       &  0.01 & 2.36 & 98.87 & 98.93 & 99.12 \\
\hline
\multirow{ 4}{*}{Ours - ODM} & \multirow{ 4}{*}{99.64} & Novelty           & 0.16 & 1.67 & 99.95 & 99.94 & 99.96 \\
                       &                         & Gaussian noise    & 0.00 & 1.76 & 99.14 & 99.46 & 97.66 \\
                       &                         & SVHN              & 0.00 & 0.96 & 99.65 & 99.41 & 99.89 \\
                       &                         & CIFAR-10          & 0.00 & 1.31 & 99.54 & 99.45 & 99.68 \\
\hline
\end{tabular}}
\end{center}
\end{table}
\section{Experimental results on additional Tsinghua splits}
Alternatively to the Tsinghua split generated with the restrictions introduced in Section 4.2, we also perform the comparison in a set of 10 random splits without applying any restriction to the partition classes.  We still discard the classes with less than 10 images per class. Table~\ref{table:tsinghua_random} shows the average performance for this set of splits with their respective standard deviation. Since the split of the classes is random,  this leads to highly similar or mirrored classes to be separated into in-distribution and out-distribution, creating situations that are very difficult to predict correctly. For instance, detecting that a turn-left traffic sign is part of the in-distribution while the turn-right traffic sign is part of the out-distribution, is very difficult in many cases. Therefore, the results from the random splits have a much lower performance, specially for the novelty case.

When comparing the metric learning based methods, ODM improves over ML for the test set that has been seen as out-distribution during training. In general, using novelty data as out-distribution makes an improvement over said test set, as well as for background and noise. However, when using background images to push the out-of-distribution further from the in-distribution class clusters in the embedding space, novelty is almost unaffected. The same happens when noise is used as out-distribution during training. This could be explained by those cases improving the embedding space for data that is initially not so far away from the in-distribution class clusters. This would change the embedding space to push further the anomalies, but would leave the novelty classes, originally much closer to the clusters, almost at the same location.

When introducing out-of-distribution samples, the behaviour on the random splits is the same as for the restricted splits: while introducing novelty helps the detection on all cases, introducing anomaly helps the detection of the same kind of anomaly.
\begin{table}
\begin{center}
\caption{Comparison between ODIN and our proposed learning strategies on a WRN-28-10 architecture, when using novelty, anomaly (background patches and Gaussian noise) as seen out-of-distribution data as well as not seen out-of-distribution. The experiments are performed on a set of 10 random splits and the metrics provided are the mean of the metrics on the individual splits $\pm$ its standard deviation.}
\label{table:tsinghua_random}
\resizebox{\textwidth}{!}{
\begin{tabular}{ccccccccc}
\hline
\noalign{\smallskip}
\multirow{ 2}{*}{Method}  & In-dist  & \multirow{ 2}{*}{Out-dist} & FPR at    & Detection    & \multirow{ 2}{*}{AUROC} & \multirow{ 2}{*}{AUPR-in} & \multirow{ 2}{*}{AUPR-out} \\
                          & accuracy &                            & 95\% TPR  & error &  &  & \\
\hline
\multirow{ 3}{*}{ODIN}    & \multirow{ 3}{*}{99.29$\pm$0.05} & Tsinghua (unseen)   & 20.85$\pm$2.28  & 12.92$\pm$1.14 & 93.50$\pm$1.05 & 93.78$\pm$1.93 & 92.41$\pm$0.73 \\
                          &                                  & Background (unseen) &  8.39$\pm$6.34 &  6.70$\pm$3.17 & 98.06$\pm$1.26 & 97.02$\pm$3.15 & 98.79$\pm$0.60 \\
                          &                                  & Noise (unseen)      &  0.03$\pm$0.43  &  2.53$\pm$0.85 & 99.67$\pm$0.34 & 99.60$\pm$0.39 & 99.74$\pm$0.41 \\
\hline
\multirow{ 3}{*}{Ours - ML} & \multirow{ 3}{*}{99.16$\pm$0.16} & Tsinghua (unseen)   & 21.05$\pm$3.25 & 13.03$\pm$1.62 & 94.18$\pm$0.92 & 94.42$\pm$1.12 & 92.75$\pm$1.08 \\
                            &                                  & Background (unseen) & 1.91$\pm$1.02 & 3.45$\pm$0.51 & 99.14$\pm$0.32 & 98.79$\pm$0.35 & 99.40$\pm$0.22 \\
                            &                                  & Noise (unseen)      &  0.30$\pm$0.96 & 2.65$\pm$0.48 & 99.27$\pm$0.36 & 99.09$\pm$0.40 & 99.43$\pm$0.35 \\
\hline
\multirow{ 3}{*}{Ours - ODM} & \multirow{ 3}{*}{99.13$\pm$0.22} & \textbf{Tsinghua} (seen)   & 16.29$\pm$4.53 & 10.65$\pm$2.26 & 96.27$\pm$0.86 & 96.78$\pm$0.93 & 95.11$\pm$1.15 \\
                             &                                  & Background (unseen)        &  0.39$\pm$1.63 &  2.71$\pm$0.31 & 99.50$\pm$0.27 & 99.30$\pm$0.31 & 99.66$\pm$0.20 \\
                             &                                  & Noise (unseen)             &  0.01$\pm$1.39 &  2.51$\pm$0.70 & 99.59$\pm$0.54 & 99.51$\pm$0.60 & 99.69$\pm$0.43 \\
\hline
\multirow{ 3}{*}{Ours - ODM} & \multirow{ 3}{*}{99.09$\pm$0.18} & Tsinghua (unseen)          & 20.36$\pm$3.63 & 12.68$\pm$1.81 & 93.47$\pm$1.55 & 93.58$\pm$2.10 & 92.00$\pm$1.74 \\
                             &                                  & \textbf{Background} (seen) &  0.01$\pm$0.03 &  2.51$\pm$0.01 & 99.97$\pm$0.02 & 99.92$\pm$0.03 & 99.98$\pm$0.01 \\
                             &                                  & Noise (unseen)             &  0.00$\pm$0.00 &  2.50$\pm$0.01 & 99.99$\pm$0.03 & 99.97$\pm$0.05 & 99.99$\pm$0.01\\
\hline
\multirow{ 3}{*}{Ours - ODM} & \multirow{ 3}{*}{99.02$\pm$2.42} & Tsinghua (unseen)          & 20.87$\pm$1.63 & 12.93$\pm$0.81 & 93.65$\pm$1.05 & 94.01$\pm$1.48 & 92.33$\pm$0.89 \\
                             &                                  & Background (unseen)       & 0.97$\pm$1.19 & 2.99$\pm$0.60 & 99.14$\pm$0.19 & 98.90$\pm$0.23 & 99.39$\pm$0.19 \\
                             &                                  & \textbf{Noise} (seen)      & 0.00$\pm$0.00  & 2.50$\pm$0.01 & 100.00$\pm$0.00 & 99.98$\pm$0.01 & 99.99$\pm$1.85\\
\hline
\end{tabular}}
\end{center}
\end{table}
\section{Embeddings on Tsinghua}
\begin{figure}
\centering
\includegraphics[width=.38\textwidth]{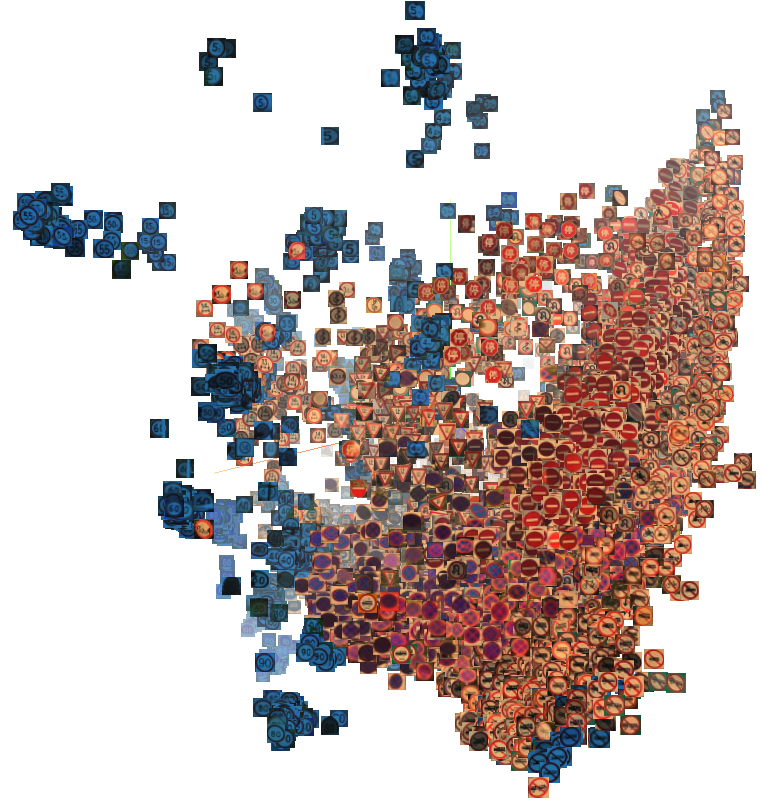}~\qquad
\includegraphics[width=.48\textwidth]{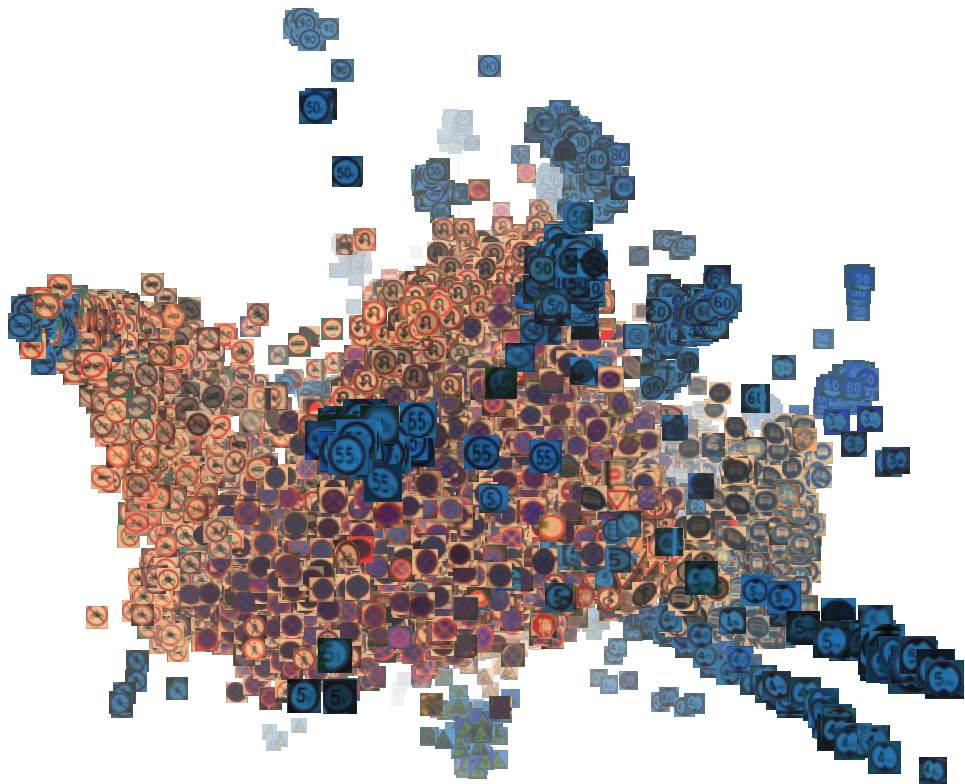}\\
\includegraphics[width=.48\textwidth]{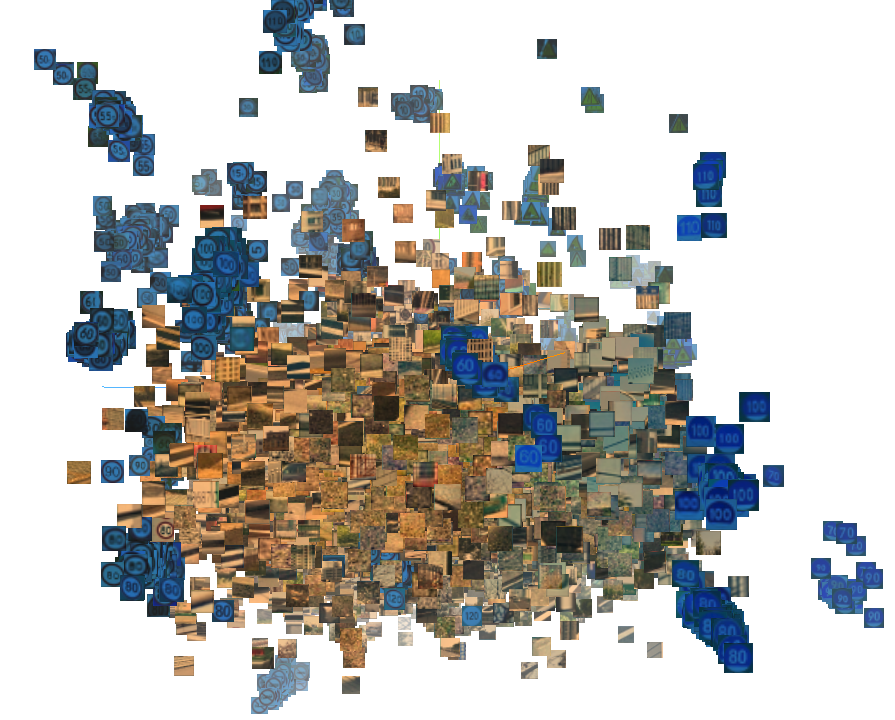}~
\includegraphics[width=.48\textwidth]{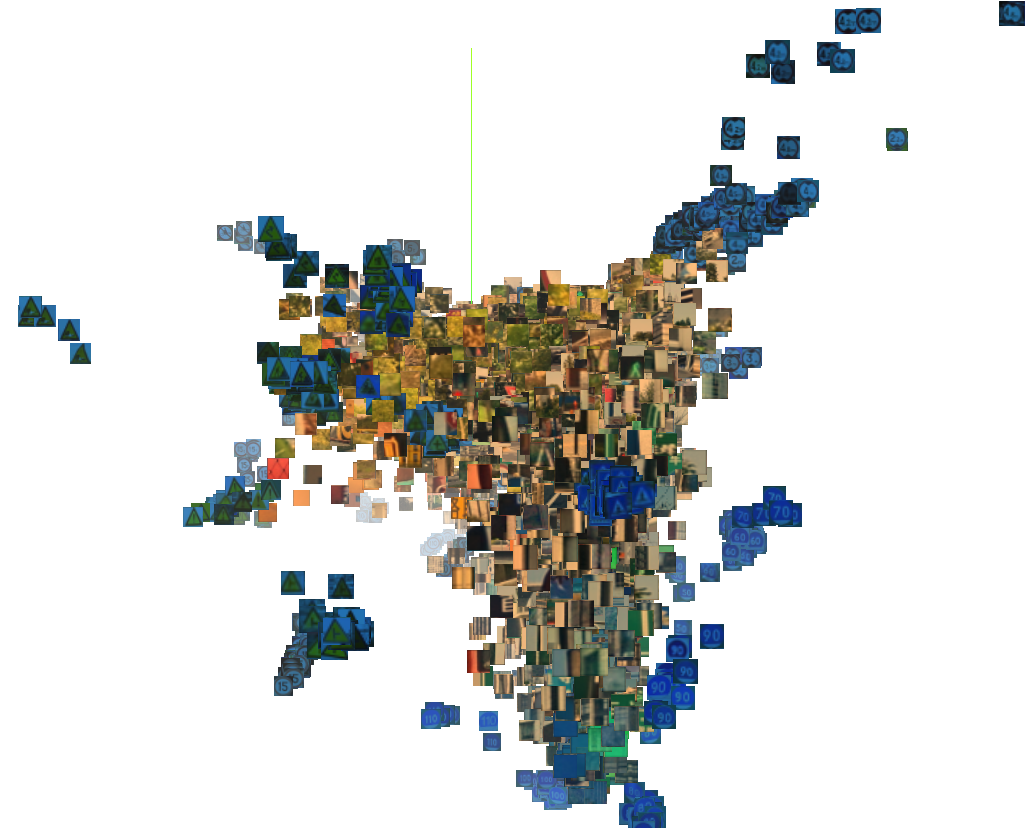}\\
\includegraphics[width=.48\textwidth]{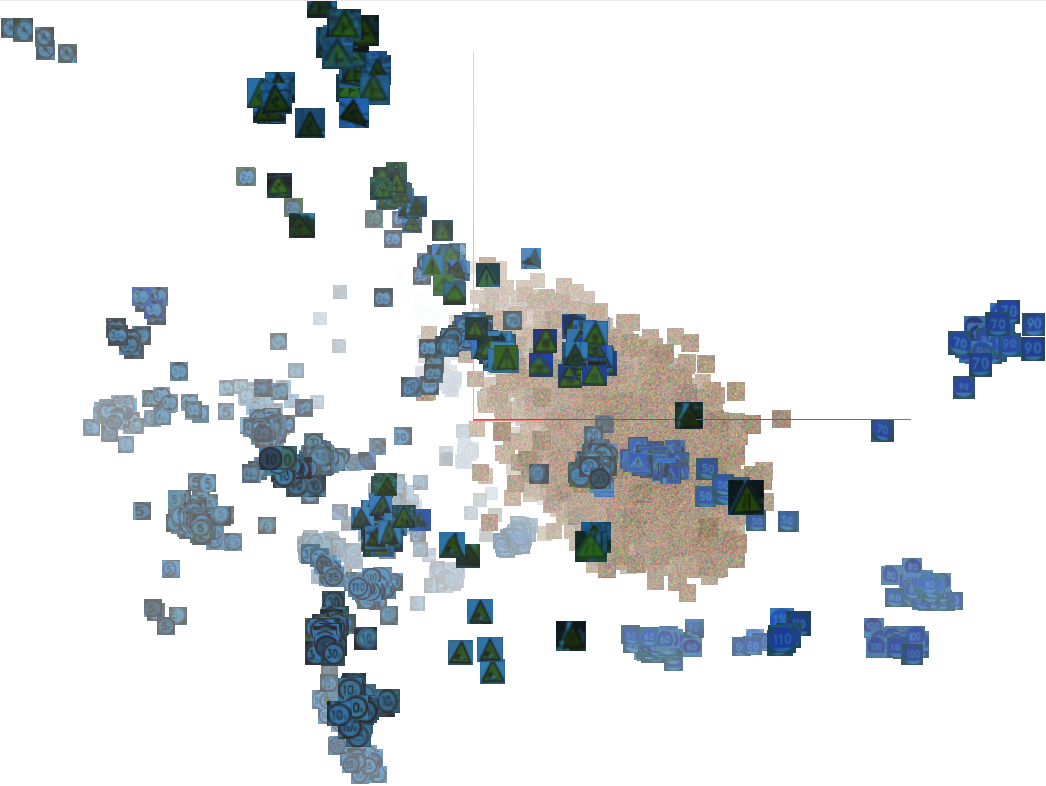}~\qquad
\includegraphics[width=.38\textwidth]{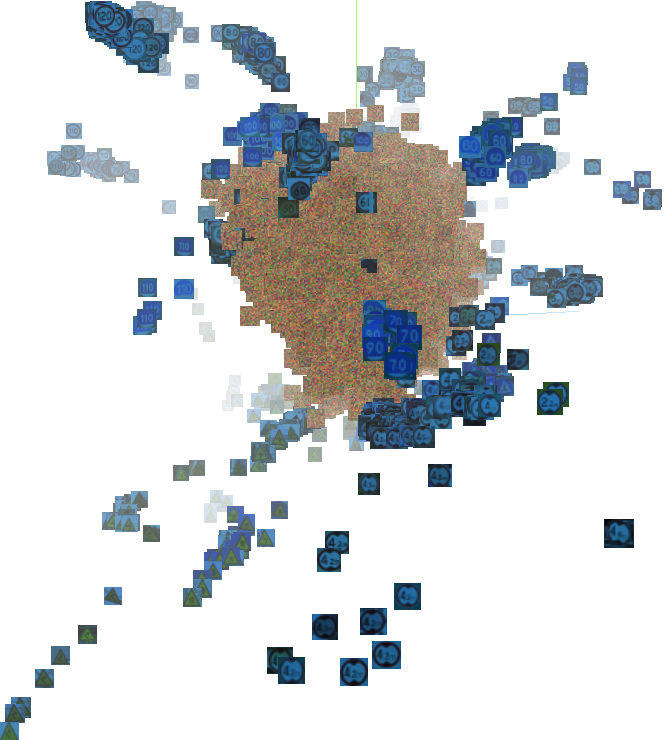}
\caption{Embedding spaces after PCA for ODM (left) and ML (right) tested for in-dist (blue shaded) and out-dist (yellow shaded). Results are for TSinghua (first row), background patches (second row) and Gaussian noise (third row). Best viewed in color.}
\label{fig:embeddingsTSinghua}
\end{figure}

Figure~\ref{fig:embeddingsTSinghua} shows the embeddings for ODM (with novelty as seen out-of-distribution) and ML after applying PCA. When using ML, the novelties are not forced to be pushed away from the in-distribution clusters so they share the embedding space in between those same in-distribution clusters. In the case of ODM, the out-of-distribution clusters are more clearly separated from the in-distribution ones.

\end{document}